\documentclass[letterpaper, 10 pt, conference]{ieeeconf}  

\IEEEoverridecommandlockouts                              

\usepackage{graphics} 
\usepackage{epsfig} 
\usepackage{amsmath} 
\usepackage{amssymb}  
\usepackage{siunitx}
\usepackage{tabularx}
\usepackage{booktabs}
\usepackage{mdframed}
\usepackage{url}
\title{\LARGE \bf
Agri-SAGE: Simulation-Grounded Multi-Agent LLM  for Context-Aware Agricultural Advisory Generation
}

\author{%
Vedant Balasubramaniam$^{1}$,
Geetha Charan$^{1}$,
Manojkumar Patil$^{1}$,
Rohit P Suresh$^{1}$,
V Priyanka$^{1}$, \\
Kodur Sai Vinay Sathvik$^{2}$
and Y. Narahari$^{1}$
\thanks{$^{1}$Indian Institute of Science, Bengaluru
{\tt\small \{vedantb, geethacharan, pmanojkumar, rohitpsuresh, priyankav, narahari\}@iisc.ac.in}}%
\thanks{$^{2}$BNM Institute of Technology, Bengaluru
{\tt\small sathvik2004@gmail.com}}%
}

\begin{document}

\maketitle
\thispagestyle{empty}
\pagestyle{empty}

\begin{abstract}
Agricultural advisory systems face a fundamental tension: static agronomic guidelines offer consistent, evidence-based recommendations, yet remain blind to in-season variability and dynamic uncertainties. Recent advisory systems powered by LLMs are liable for a different risk of generating recommendations that are agronomically credible but physiologically unconvincing.
Agri-SAGE is a closed-loop framework designed to resolve the above two limitations by integrating retrieval-grounded multi-agent LLM reasoning with APSIM-based biophysical simulation, to generate and validate agronomic advisories.. To assess this framework, we evaluate three reasoning approaches, namely {\em Plan-and-Solve\/}, {\em Tree of Thoughts\/}, and {\em Reflexion\/}, over a 10-year retrospective analysis. All three significantly outperform static PoP (Package-of-Practice) baselines, with Tree of Thoughts achieving impressive peak yields. At the same time, Reflexion achieves comparable agronomic outcomes at substantially lower computational cost by leveraging cross-seasonal episodic memory.

\end{abstract}

\section{INTRODUCTION}

Agriculture remains the cornerstone of food security and rural livelihoods across much of the developing world. In India, where a substantial share of the population depends on farming for subsistence and income, the quality of agricultural guidance available to farmers has profound implications for productivity and economic stability.

Traditionally, agricultural advisories in India are delivered through \textit{Packages of Practices} (PoPs). 
which contain agronomic guidelines derived from extensive field trials and agronomic research. These recommendations typically include land preparation methods, fertilizer schedules, irrigation practices, and pest management strategies.

While scientifically grounded, PoPs are inherently static as they are issued before the cropping season at the agro-climatic zone level and updated infrequently. As a result, they cannot adapt to season variability such as weather shocks, pest outbreaks, or localized soil heterogeneity. Consequently, farmers receive recommendations that may not be suitable to their specific field conditions.

To partially address this limitation, agricultural advisory systems also provide \textit{In-Season Advisories}. These are dynamic, time-sensitive recommendations generated during the cropping season based on real-time environmental signals, including weather forecasts, temperature patterns, pest incidence, and crop phenological stages. In India, Agro-Meteorological Field Units (AMFUs) are responsible for generating such advisories. However, these advisories remain largely manual and expert-driven, limiting scalability and the ability to personalize recommendations.

Recent advances in LLMs have enabled agricultural advisory systems capable of synthesizing large agronomic knowledge bases and interacting with farmers through natural language interfaces. While these systems significantly improve accessibility and knowledge dissemination, they remain constrained as they may generate plausible but incorrect recommendations. 

To address these limitations, we introduce Agri-SAGE, a novel framework that reconceives the traditional Package of Practices as a dynamic, context-aware agronomic plan. 
At its core, Agri-SAGE uses a multi-agent LLM architecture, with specialized agents to generate, evaluate, and refine agricultural recommendations. By combining agronomic knowledge retrieval with real-time environmental inputs, the system produces season-long management strategies that are coherent and adaptive.

A central challenge in AI-driven agronomic advice is ensuring that recommendations remain physically plausible and safe to act on. To address this, Agri-SAGE validates all generated advisory plans using the Agricultural Production Systems sIMulator (APSIM) ~\cite{holzworth2014apsim}, a widely adopted process-based crop simulation model. Unlike prior agricultural advisory systems that rely solely on textual reasoning, Agri-SAGE grounds every recommendation in crop physiology through APSIM simulation, ensuring that generated advisories remain biophysically feasible.

\noindent\textbf{Contributions}
\begin{itemize}
    \item We propose a novel, closed-loop autonomous AI Agronomist framework that couples LLMs with the APSIM biophysical crop simulator, successfully grounding generative semantic reasoning 

    \item We conduct a rigorous 10-year comparative analysis of advanced reasoning methodologies, namely \textit{Plan-and-Solve (PS)}, \textit{Tree of Thoughts (ToT)}, and \textit{Reflexion}. We demonstrate that all these approaches significantly outperform static PoP baselines with \textit{Tree of Thoughts} achieving the highest yield. 

    \item We observe that the proposed method discovers significantly better agronomic practices than static PoP. In Mandya region, when maize is simulated under local conditions the optimized fertilizer, irrigation splits along with other operations resulted in better outcome.
\end{itemize}



\section{RELATED WORK}
\subsection{LLM-Based Agricultural Advisory Systems}

Recent work has explored the application of LLMs for automated agricultural advisory through retrieval-augmented generation (RAG) and domain adaptation. Farmer.Chat~\cite{singh2024farmer} demonstrates large-scale deployment of conversational agricultural advisory using curated agronomic knowledge bases combined with weather APIs. Similar systems such as ShizishanGPT~\cite{yang2024shizishangpt}, AgriGPT~\cite{yang2025agrigpt}, and AgriRegion~\cite{fanuel2025agriregion} extend this paradigm through techniques including knowledge graphs, region-aware retrieval, and domain-specific fine-tuning.

These approaches significantly improve factual grounding compared to generic LLMs and reduce hallucinations through structured retrieval pipelines~\cite{singh2026fine}. However, they remain fundamentally limited by the knowledge available within their underlying corpora. Consequently, they cannot reason about agronomic scenarios that are absent from the retrieved documents.

Several studies highlight the reliability challenges associated with corpus-grounded advisory systems. Evaluations of GPT-based systems for pest management show that performance improves substantially only when explicit domain context is injected~\cite{yang2024gpt}. Broader analyses demonstrate that LLMs may generate confident but incorrect agronomic recommendations when operating outside their training distribution~\cite{de2024large}. Multi-agent frameworks such as AgroAskAI attempt to mitigate these issues by introducing specialized agents and reviewer modules that critique generated responses~\cite{cantonjos2025agroaskai}. 

\subsection{Simulation Models and Multi-Agent Systems}

Process-based crop simulation models provide a complementary mechanism for evaluating management decisions. Systems such as APSIM NextGen engine simulate crop growth, water balance, and nutrient dynamics at daily resolution and are widely used for agronomic scenario analysis.

Recent studies have begun exploring the integration of language models with simulation environments. Wu et al.~\cite{wu2024new} demonstrate that language models can reason over DSSAT simulator outputs for crop management optimization. MCP-SIM~\cite{park2026self} introduces a self-correcting multi-agent framework that transforms underspecified prompts into validated simulations through iterative plan–act–reflect–revise cycles, improving convergence efficiency across multiple task settings.

More broadly, multi-agent LLM frameworks have demonstrated effectiveness across complex reasoning tasks. Jia et al.~\cite{jia2025enhancing} proposed a feedback-driven multi-agent framework that combines enhanced retrieval, reasoning modules, and environmental actions, achieving strong performance in power system optimization tasks. Xia et al.~\cite{xia2024llm} presented specialized LLM agents for automated parametrization of simulation models in digital twins. Similarly, Reflexion~\cite{shinn2023reflexion} introduces episodic memory-based feedback mechanisms that enable language agents to iteratively refine solutions without updating model weights. CodeSim~\cite{islam2025codesim} extends this paradigm to program synthesis through internal debugging via simulated input–output execution.



Existing systems either rely solely on textual reasoning without physiological validation or optimize isolated decisions without producing complete season-long plans. Agri-SAGE addresses both gaps by combining LLM-based retrieval and reasoning with APSIM simulation, ensuring that full-season agronomic advisories are grounded in crop physiology rather than solely in text.

\section{METHODOLOGY}
\begin{figure*}[tb]
    \centering
\includegraphics[width=0.9\linewidth]{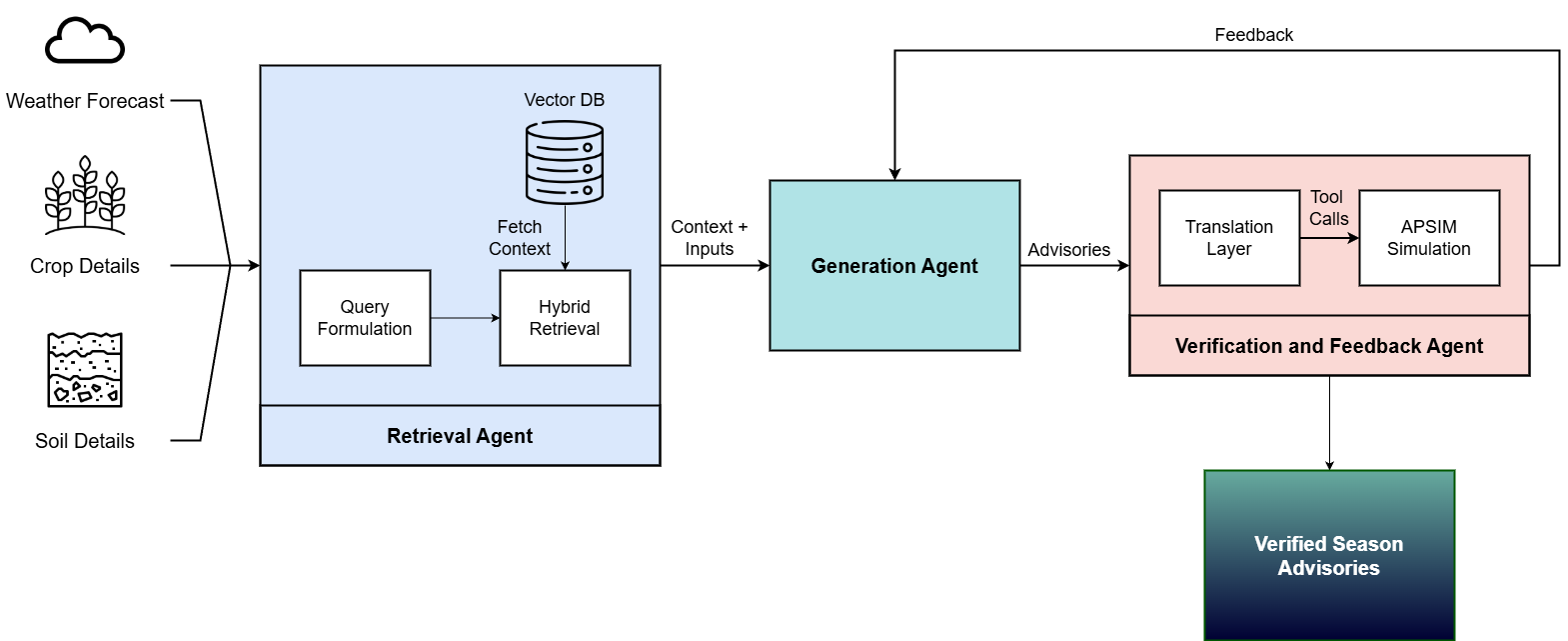}
    \caption{Architecture of Agri-SAGE}
    \label{fig:agri-sage}
\end{figure*}

Figure \ref{fig:agri-sage} shows the architecture of Agri-SAGE. 
It comprises three primary modules: 1) Retrieval Agent, 2) Generation Agent, and 3) Verification and Feedback Agent.

\subsection{The Retrieval Agent}
The framework uses a Retrieval Agent to ground LLM outputs in a curated knowledge base of regional manuscripts and Package of Practices (PoPs). By seeding the action context with PoPs from local agricultural universities, it anchors recommendations to crop inputs, fertilizer types, and operational methods that farmers in the area can realistically access and afford.


To address the high dimensionality of agricultural texts, we implemented semantic-aware hierarchical chunking in combination with hybrid dense–sparse retrieval mechanisms. 
The Retrieval Agent dynamically processes environmental inputs, such as real-time weather forecasts, soil profiles, crop phenology, and solar radiation, and translates them into optimized vector search queries. This ensures that the Generation Agent receives highly relevant, context-specific agronomic literature prior to formulating its advisory outputs.

\begin{figure*}[tb]
    \centering
\includegraphics[width=0.9\linewidth]{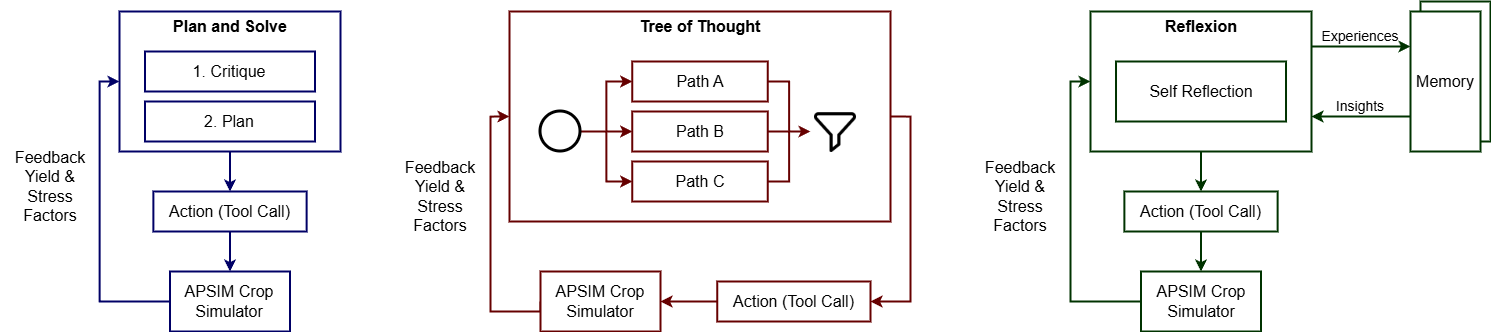}
    \caption{Reasoning Methodologies: Plan and Solve (left), Tree of Thought (middle), and Reflexion (right).}
    \label{fig:agri-sage-reasoning}
\end{figure*}
\subsection{The Generation Agent}
The Generation Agent serves as the central decision-making component of the proposed framework. It synthesizes the localized agronomic context retrieved by the Retrieval Agent with the real-time environmental state to formulate detailed, actionable agricultural advisories.

We formalize the agronomic advisory task as an iterative optimization problem where the APSIM simulator serves as a deterministic environment providing an observation after each iteration. The LLM generates an optimized action using the observations from APSIM. We evaluate three distinct reasoning methodologies namely, Plan-and-Solve (PS), Tree of Thoughts (ToT), and Reflexion illustrated in Figure \ref{fig:agri-sage-reasoning}.

\noindent\textbf{Plan-and-Solve (PS)}

We adapt the PS prompting strategy \cite{wang2023plan} into a closed-loop biophysical framework. At each iteration, the agent receives the prior observation and processes it sequentially: first generating a textual critique diagnosing the specific biophysical failures of the previous action, then generating a revised plan, and finally executing the new action.



The deterministic feedback from the APSIM simulator, along with baseline yields, guides the LLM  to autonomously self-correct. Through this mechanism, the agent iteratively converges on optimized and improved agronomic actions.

\noindent\textbf{Tree of Thoughts (ToT)}

Following Yao et al. \cite{yao2023tree}, ToT is implemented as an anticipatory lookahead search. Before executing any action in the simulator, the agent generates {\em{k=3}} distinct, mutually exclusive agronomic intervention pathways. An internal LLM evaluation explicitly critiques the biophysical pros and cons of each path against the current weather forecast. The agent selects the highest-scoring path to translate into the final action. Only this advisory is sent to the APSIM simulator. By explicitly forcing the generation of three divergent branches prior to execution, ToT prevents the agent from aggressively optimizing a fundamentally flawed initial strategy.


\noindent\textbf{Reflexion}

Our Reflexion implementation follows Shinn et al. \cite{shinn2023reflexion}, extending the PS framework by integrating a persistent episodic memory module. In this context, episodic memory is defined as a textual database that stores distilled heuristic rules derived from past simulation years (e.g., logging that a mid-June sowing date historically fails during drought years). After each simulated growing season, the agent generates a reflection diagnosing the season's macro-level success or failure, which is appended to memory. In subsequent years, the policy is conditioned on this historical context. By caching biophysical dynamics into a textual memory bank, Reflexion enables zero-shot inference-time adaptation across diverse weather cycles.

\subsection{The Verification and Feedback Agent}

Within the Verification and Feedback Agent, a translation layer converts the output of the Generation Agent into structured, executable tool calls for APSIM  which include setting the sowing date, type of fertilizer (ex Urea, DAP, etc) and its amount and date of application, amount of irrigation(mm) and date of application, and more operations like surface mulch (kg/ha) and tillage events.

APSIM then simulates the full crop life cycle under the prescribed advisories. While the APSIM engine dynamically computes various biophysical variables (e.g., Leaf Area Index, total above-ground biomass, and carbon dynamics), the Feedback Agent isolates and extracts three primary metrics to form the feedback:

\begin{enumerate}
    \item \textbf{Grain Yield (\si{kg/ha})}: The final quantitative output of the crop, serving as the primary economic objective function for optimization.
    \item \textbf{Water Stress Factor}: A daily physiological metric indicating hydration sufficiency.
    \item \textbf{Nitrogen Stress Factor}: A daily physiological metric indicating nutrient deficiency.
\end{enumerate}

We selected these three metrics because Grain Yield serves as the ultimate benchmark of agronomic success, while the stress factors map directly to the primary actionable components: irrigation scheduling and fertilizer application.

By exposing the underlying physical drivers of crop failure or success (e.g., nitrogen leaching due to premature application before heavy rainfall), the LLM can iteratively critique and refine its previous advisories to maximize yield. 




\section{DATASETS AND EXPERIMENTAL SETUP}

\subsection{Weather, Region, and Crop Selection}

\begin{itemize}
    \item \textbf{Region and Climate:} We selected Mandya, Karnataka, India, a semi-arid zone known for high year-to-year weather variability. Between 2015 and 2024, this region experienced severe droughts and extreme monsoons, providing a rigorous testing ground to evaluate how well LLM agents adapt to climate stress.
    \item \textbf{Weather Data:} 10 years of continuous daily weather data (2015–2024) were sourced from the ERA5 dataset\footnote{\url{https://climate.copernicus.eu}}, including solar radiation, temperature, and precipitation. 
    \item \textbf{Soil Profile:} The simulator was parameterized using Mandya red sandy soil (Alfisol), defining the baseline water-holding capacity, bulk density, and organic carbon content.
\end{itemize}


We restrict our experiments to a single crop: Maize (\textit{Zea mays}). Although paddy (rice) is the primary staple crop in Mandya region, localized paddy models in the APSIM framework are currently limited. Maize, one of the predominantly cultivated crops in the region, is economically critical and requires precise management to maintain yields during weather extremes. Using a single crop establishes a controlled baseline, allowing us to accurately measure the LLM agents' reasoning capabilities without the added complexity of multi-crop dynamics. {\em Though we focus on maize,  the architecture is crop-agnostic and works for any crop supported by APSIM\/}.

\subsection{Agronomic Knowledge Base}



The retrieval corpus comprised approximately 1,000 open-access research papers on maize cultivation, nitrogen use efficiency, and drought mitigation, scraped from Semantic Scholar, and supplemented by official Karnataka agricultural extension materials covering recommended sowing windows and cultivar selection.






\subsection{Baseline Formulation}
We establish a benchmark using the ``Gold Standard" Package of Practices (PoP) for the region of study, which represents the static advisory schedule followed by local farmers. This PoP was translated into fixed APSIM tool calls, specifying predetermined sowing dates and fertilizer splits, and simulated across the 10-year weather dataset to generate annual baseline yields. These yields are provided to the Generation Agent as an optimization threshold against which the proposed framework was evaluated.

\subsection{Experimental Setup}
Experiments were conducted using a 4-bit quantized DeepSeek-R1 model with the generation temperature set to $0.3$. This low temperature ensures that the model strictly adheres to JSON tool-calling syntax and deterministic agronomic logic, while retaining sufficient stochasticity to explore valid biophysical interventions without hallucinating unrealistic constraints.

To account for the inherent stochasticity of LLM generation, each configuration (Architecture $\times$ Feedback Depth $\times$ Year) was executed in 10 independent runs. This sample size captures the distribution of exploratory actions and stabilizes statistical variance, balancing statistical robustness with the computational cost of APSIM simulations. The metrics are reported as the mean across runs to ensure robust comparative baselines.

To quantify the Simulator-in-the-Loop impact, we performed an ablation study over feedback depths of $N \in \{0, 2, 3, 4\}$. The 0-loop configuration establishes a strict zero-shot baseline relying solely on RAG context. The multi-shot configurations permit up to 4 iterative feedback loops. The upper bound was deliberately capped at 4 iterations because empirical data indicates agents reach a biophysical optimization plateau within 3 to 4 loops.

\section{EXPERIMENTAL RESULTS}

\subsection{Quantitative Analysis}
\begin{figure}[h]
    \centering
    \includegraphics[width=\linewidth]{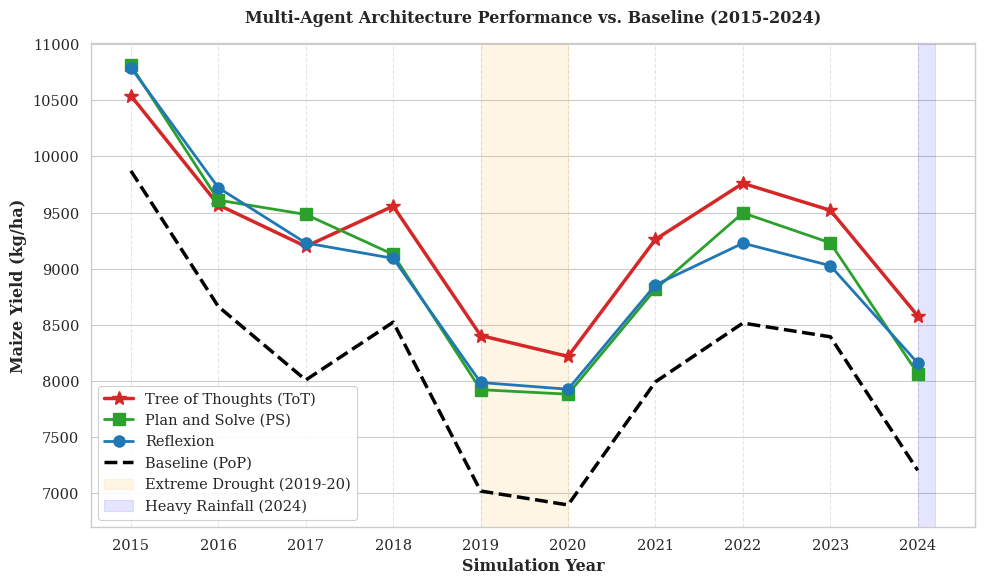}
    \caption{Yield Performance of Plan-and-Solve, Reflexion, and Tree of Thoughts versus baseline over 10‑year horizon (mean over 10 runs)}.
    \label{fig:quant}
\end{figure}

As illustrated in Figure \ref{fig:quant}, all three reasoning architectures consistently outperformed the static baseline across the entire decade. The baseline achieved a 10-year average yield of 8,110 kg/ha, but struggled significantly during extreme weather events, such as the 2019 drought (7,022 kg/ha) and the 2024 heavy monsoons (7,206 kg/ha). 


Both Plan-and-Solve and Reflexion achieved similar yield ceilings, averaging 9,045 kg/ha (+935 kg/ha) and 9,002 kg/ha (+892 kg/ha), respectively. Plan-and-Solve's gains stem from its closed-loop critique-and-revise structure. By first generating a textual diagnosis of biophysical failures before issuing each new action, the agent avoids compounding earlier errors across iterations. Reflexion reaches a comparable ceiling through its persistent episodic memory bank, accumulating heuristic rules across simulated growing seasons. However, both methods ultimately converge on similar performance limits due to their partly similar methodologies.

In contrast, Tree of Thoughts (ToT) consistently outperformed all other methods by achieving a 10-year average maximum yield of 9,262 kg/ha (+1,152 kg/ha over the baseline). ToT's superiority arises from its anticipatory lookahead design. Before any action reaches the APSIM simulator, the agent generates three mutually exclusive agronomic intervention pathways and explicitly critiques each against the current weather forecast. By forcing deliberate divergence prior to execution, ToT avoids the failure mode common to reactive strategies. This structural advantage is most pronounced during severe climatological stress; for instance, during the 2019 drought and the 2024 floods, the ToT agent outperformed the baseline by 1,384 kg/ha and 1,375 kg/ha, respectively.


\subsection{Convergence Dynamics}
\begin{figure}[h]
    \centering
    \includegraphics[width=\linewidth]{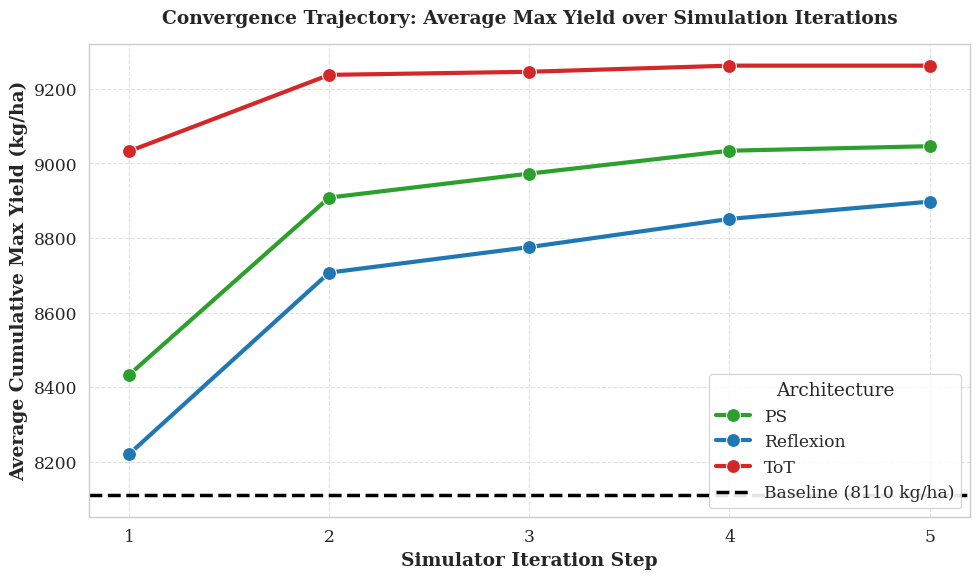}
    \caption{Convergence behavior under increasing feedback iterations.}
    \label{fig:ablation}
\end{figure}
To understand the computational efficiency of each reasoning methodology, we analyzed intra-episode convergence trajectories, tracking the average maximum yield at each iterative feedback step (Figure \ref{fig:ablation}). 

We observe that while the yields of advisories generated using PS increase significantly in the initial iterations, they flatten out after 3 to 4 feedback loops as the agent exhausts the low-hanging corrective gains. ToT often achieves optimal yield in Iteration 1 due to its tree-like exploration strategy. Similarly, Reflexion's episodic memory bank enables faster convergence. This establishes that forcing these agents beyond 3 iterations yields diminishing returns. We observe that while ToT achieves peak yields, it consumes {\em k} times more tokens per decision stage because it explores {\em k} parallel paths simultaneously. In contrast, Reflexion and PS incrementally refine a single path, making them superior in execution efficiency.


\subsection{Qualitative Analysis}

\begin{table*}[!ht]
\centering
\caption{Qualitative Comparison of Agent Strategies During the 2019 Drought}
\label{tab:qualitative_comparison}
\begin{tabular}{p{3cm} p{2.5cm} p{3cm} p{7cm}}
\toprule
\textbf{Architecture} & \textbf{Yield (kg/ha)} & \textbf{Reasoning Style} & \textbf{Key Strategy Used to Beat the Baseline} \\
\midrule
\textbf{Baseline (PoP)} & 7,022 & Static Rules & Followed a fixed calendar and suffered water stress. \\
\midrule
\textbf{Iterative Plan-and-Solve} & 8,168 & Reactive (Trial \& Error) & Used simulator feedback to apply mulching and switched to dry-soil fertilizers. \\
\midrule
\textbf{Reflexion} & 7,988 & Proactive (Memory) & Since it had never experienced drought, it had to rely on the simulator's feedback.\\
\midrule
\textbf{Tree of Thoughts} & \textbf{8,825} & Anticipatory (Lookahead) & Evaluated various paths and decided to shift the sowing date a month early to dodge the drought entirely. \\
\bottomrule
\end{tabular}
\end{table*}

To understand exactly how the AI agents achieved higher yields than the static heuristics, we analyzed their decision-making during the 2019 simulation. This year featured a severe late-season drought, making it a perfect case study to see how each architecture handles extreme weather.

\subsubsection{Baseline}
In 2019, the PoP dictated mid-June as the sowing date. Because this date is hard-coded, the maize crop entered its most sensitive reproductive stage precisely when the late-season drought was at its worst, resulting in severe water stress and a low yield of 7,022 kg/ha.

\subsubsection{Plan-and-Solve (PS)}
Plan-and-Solve starts each year with a blank slate. On its first attempt, it followed standard timing rules and suffered the same drought penalties as the baseline. However, when PS received high water stress as feedback from the APSIM simulator. the agent's \texttt{<critique>} module diagnosed the problem. 
In its next iteration, the agent actively changed its plan: it added a thick layer of surface mulch to prevent water from evaporating, and it switched its fertilizer to Nitrate, which dissolves easily even in dry soil. By responding to the simulator's feedback, PS successfully recovered from failure and reached 8,168 kg/ha.

\subsubsection{Tree of Thoughts (ToT)}

The Tree of Thoughts (ToT) agent achieved the best result (8,825 kg/ha) by evaluating different strategies before making a final decision. The ToT agent generated three possible paths by comparing a ``Conservative \& Early Sowing'' path against a ``Delayed Sowing'' path. Then it realized that planting in mid-June would ruin the crop. 

Instead of trying to fight the drought, the ToT agent simply avoided it. It shifted the sowing date a full month earlier to May 15. Because of this early start, the crop finished growing before the extreme heat arrived. ToT caught the timing error that the baseline missed, requiring only a single attempt to find the optimal solution.

\subsubsection{Reflexion}

While ToT looked ahead, Reflexion looked back. Because the Reflexion agent stores text-based summaries of past farming seasons, it was better able to anticipate and prepare for drought conditions than any other agent. In 2019, it proactively applied 5,000 kg/ha of organic manure and scheduled a deep tillage event. Mixing heavy manure deep into the soil acts like a sponge, significantly increasing the soil's water-holding capacity. The baseline missed this soil-preparation step entirely. By leveraging experience, Reflexion achieved a high yield of 7,988 kg/ha on its first try without running multiple simulator iterations. One of our observations in Reflexion was that, as the years progressed, yields in the first iteration increased, with subsequent loops yielding only minimal gains.

In the early years (2015–2016), Reflexion underwent multiple iterations as it populated its episodic memory bank. By 2017 and 2018, it successfully collapsed its simulator reliance to a single iteration. However, during the severe drought anomaly of 2019 and 2020, Reflexion's performance momentarily deteriorated, spiking to 4 iterations. 

\section{CONCLUSION AND FUTURE WORK}
We presented Agri-SAGE, a closed-loop multi-agent framework that couples LLM-based reasoning with APSIM biophysical crop simulation to generate agronomically validated, context-aware agricultural advisories. Across the 10-year retrospective evaluation, Agri-SAGE consistently outperforms the static Package-of-Practice baseline under diverse climatic conditions. Three key insights emerge from our experiments. First, all reasoning-based methods substantially improve crop yield compared to static agronomic schedules, demonstrating the effectiveness of simulation-grounded adaptive advisory generation. Second, Tree of Thoughts consistently achieves the highest yield improvements by exploring multiple agronomic intervention pathways before committing to an action. Third, Reflexion captures cross-seasonal agronomic insights via episodic memory, achieving competitive yields with fewer simulator interactions. Together, these results demonstrate that coupling LLM reasoning with crop simulation can transform the state of the art in agronomic advisories.  

Several directions remain open for future work. The optimization objective should be extended beyond yield to incorporate input costs, profit margins, and sustainability metrics such as nitrogen leaching and carbon footprint. The effect of real-time weather forecasts also needs to be analyzed. Additionally, APSIM predictions serve as a proxy for agronomic outcomes, and field validation remains an important next step given real-world variability in soil and microclimate. Finally, evaluating the framework with agronomically fine-tuned model variants, using synthetic APSIM-generated data as a training signal, could unlock stronger reasoning capabilities for novel climate scenarios outside historical distributions.



\addtolength{\textheight}{-12cm}   



\section*{ACKNOWLEDGMENT}
The authors thank Prof. Soma Dhawala for his valuable suggestions and agronomic insights. The authors also thank Dr. Richard Magala for his helpful discussions regarding the apsimNGpy framework.

The authors acknowledge the use of various AI tools and assistants, such as GPT and Claude, to enhance the writing quality and provide summaries of source materials. However, the authors diligently verified all information provided by these AI tools to ensure accuracy and factual integrity.

\bibliographystyle{IEEEtran}
\bibliography{references}

\end{document}